\documentclass[conference]{IEEEtran}
\IEEEoverridecommandlockouts

\newcommand{\IEEEAcceptedNotice}{%
To appear in the Proceedings of the 22nd IEEE International Conference
on Distributed Computing in Smart Systems and the Internet of Things
(DCOSS-IoT 2026), UrbCom Workshop.
\copyright~2026 IEEE. Personal use of this material is permitted.  Permission from IEEE must be obtained for all other uses, in any current or future media, including reprinting/republishing this material for advertising or promotional purposes, creating new collective works, for resale or redistribution to servers or lists, or reuse of any copyrighted component of this work in other works.
}

\makeatletter
\def\ps@IEEEtitlepagestyle{%
  \def\@oddfoot{%
    \parbox[b]{0.96\textwidth}{%
      \centering
      \scriptsize
      \IEEEAcceptedNotice
    }%
  }%
  \def\@evenfoot{}%
}
\makeatother

\usepackage{cite}
\usepackage{amsmath,amssymb,amsfonts}
\usepackage{algorithmic}
\usepackage{graphicx}
\usepackage{textcomp}
\usepackage{xcolor}
\usepackage{hyperref}
\usepackage{float}
\usepackage{array}
\usepackage[linesnumbered,ruled,vlined]{algorithm2e}
\usepackage{enumitem}

\def\BibTeX{{\rm B\kern-.05em{\sc i\kern-.025em b}\kern-.08em
    T\kern-.1667em\lower.7ex\hbox{E}\kern-.125emX}}
\begin{document}

\title{Neural Architecture Search for Traffic Prediction: A Survey of Methods, Challenges, and Future Directions}

\author{\IEEEauthorblockN{Truong Giang Vu}
\IEEEauthorblockA{\textit{Faculty of Business \& IT} \\
\textit{Ontario Tech University}\\
Oshawa, Canada \\
truonggiang.vu@ontariotechu.net}
\and
\IEEEauthorblockN{Li Yang}
\IEEEauthorblockA{\textit{Faculty of Business \& IT} \\
\textit{Ontario Tech University}\\
Oshawa, Canada \\
li.yang@ontariotechu.ca}
\and
\IEEEauthorblockN{Richard W. Pazzi}
\IEEEauthorblockA{\textit{Faculty of Business \& IT} \\
\textit{Ontario Tech University}\\
Oshawa, Canada \\
richard.pazzi@ontariotechu.ca}

}

\maketitle

\bstctlcite{IEEEexample:BSTcontrol}

\begin{abstract}
Traffic prediction is a core task in intelligent transportation systems, supporting applications such as adaptive signal control, route guidance, and ride-hailing dispatch. Deep learning models, including graph convolutional networks, recurrent networks, and Transformers, achieve strong results on standard benchmarks, but their architectures are designed by hand, requiring significant expert effort and producing models that often generalize poorly across cities and datasets. Neural Architecture Search (NAS) offers a systematic alternative to manual design. It automates the search over candidate architectures of deep learning models, finding designs that match the spatial-temporal structure of traffic data without manual trial and error. This survey reviews NAS methods applied to traffic prediction, organized by search strategy: gradient-based methods, evolutionary methods, and one-shot weight-sharing methods. For each category, we analyze how the search space is designed to cover spatial and temporal traffic operators, and how the search strategy balances cost against architecture quality. We also discuss open challenges, computational scalability to large road networks, manual search space design, cross-city generalization, dynamic graph structure, and the open question of NAS for spatial-temporal foundation models, and identify directions for future research.
\end{abstract}


\section{Introduction}\label{sec:introduction}
Traffic prediction plays an important role in Intelligent Transportation Systems (ITS).  It forecasts future traffic states such as speed, flow, occupancy, and travel demand. Accurate predictions support many real-world applications, which include traffic signal control and congestion management \cite{chavhan2020prediction}, route planning \cite{dai2020hstgcn}, and ride-hailing demand forecasting \cite{yao2018deepmvst}. As cities grow and road networks become more complex, prediction errors could have a greater impact. A single missed congestion event can cause delays across an entire city. Consequently, there is growing interest in data-driven methods that can use large amounts of sensor, Global Positioning System (GPS), and vehicle data collected in modern cities \cite{liu2023largest}.

Classical statistical methods such as Autoregressive Integrated Moving Average (ARIMA) assume linear and stationary relationships, which limits their ability to capture the nonlinear, time-varying dynamics of real traffic data \cite{li2018dcrnn}. Machine learning (ML) overcomes these limits by learning directly from data and becomes the leading approach for traffic prediction. Deep learning models, including recurrent neural networks (RNN/LSTM) for temporal patterns \cite{li2018dcrnn}, convolutional neural networks (CNN) for grid-based spatial data \cite{zhang2017stresnet}, graph convolutional networks (GCN) for road network structure \cite{yu2018stgcn}, and Transformers for long-range dependencies \cite{jiang2023pdformer}, have shown a strong ability to learn complex spatial-temporal patterns from raw data and achieve state-of-the-art results on standard benchmarks such as METR-LA, PeMS-BAY, and PeMS04/08 datasets \cite{shao2022decoupled}.

However, building good deep learning models for traffic prediction is still difficult. Traffic data has a spatial-temporal structure. Road sensors are placed in irregular networks, patterns vary between different spatial scales, and traffic behavior changes over time \cite{shao2022decoupled}. Designing a model requires many decisions, which graph convolution to use, how to combine spatial and temporal components, and how to structure the overall network, and models that work well on one city often perform poorly on another, as their architecture is typically fixed to a specific dataset \cite{pan2021autostg}. They also need significant effort to adapt when applied to new tasks or datasets \cite{wu2022autocts}. This complexity makes manual architecture design not only time-consuming but also unreliable, which motivates the need for automated approaches \cite{lyu2025autostf}.

Automated Machine Learning (AutoML) reduces the manual effort in designing and tuning ML pipelines. It does this by automating decisions that would typically require expert knowledge. Neural Architecture Search (NAS) is the part of AutoML focused on automating model architecture design of deep learning models. It defines a search space of candidate operations and uses a search strategy to find good architectures efficiently \cite{liu2019darts}. NAS methods for traffic prediction can be grouped by their search strategy. Gradient-based methods, including Automated Spatio-Temporal (AutoST) \cite{li2020autost}, Automated Spatio-Temporal Graph (AutoSTG) \cite{pan2021autostg}, AutoSTG+ \cite{ke2023autostgplus}, Automated Spatio-Temporal Learning (AutoSTL) \cite{zhang2023autostl}, and Neural Architecture Search for Spatio-Temporal (NASST) \cite{zhang2023nasst}, use differentiable relaxation to jointly search over spatial and temporal operations. Evolutionary methods, such as those by Klosa et al. \cite{klosa2022evolutionary} and Low-cost Evolutionary Neural Architecture Search (LENAS) \cite{klosa2023lenas}, apply population-based search to find compact architectures at low cost. One-shot and weight-sharing methods, Automated Correlated Time Series (AutoCTS) \cite{wu2022autocts}, AutoCTS+ \cite{wu2023autoctsplus}, and Automated Spatio-Temporal Forecasting (AutoSTF) \cite{lyu2025autostf}, train a single supernet shared across all candidate architectures, greatly reducing search time. Across all these categories, automatically found architectures are often reported to match or outperform hand-designed models, making NAS a promising direction for traffic prediction \cite{lyu2025autostf}.

This survey provides a focused review of NAS methods applied to traffic prediction. We make three main contributions. First, we organize existing work into a clear taxonomy based on search space design and search strategy. Second, we connect each NAS design choice to the specific challenges of spatial-temporal traffic data. Third, we identify open problems of using NAS for traffic prediction and suggest future research directions for researchers and practitioners working in this area.

Several surveys review adjacent territory: NAS methods \cite{elsken2019nas, white2023nas1000}, graph neural networks for traffic \cite{jiang2022gnntrafic, jin2023stgnnsurvey}, and AutoML for traffic prediction \cite{khatiriolyaee2026automl}, but none focuses specifically on how NAS search strategies and search spaces are designed for spatio-temporal traffic data. Papers included in this review were identified by searching Google Scholar, IEEE Xplore, and ACM Digital Library using the keywords \textit{neural architecture search}, \textit{traffic prediction}, \textit{spatio-temporal forecasting}, and \textit{automated machine learning}, covering work published between 2019 and 2025; we included papers that directly apply NAS or automated architecture design to traffic or spatio-temporal benchmarks.

The rest of this survey is structured as follows. Section II defines the traffic prediction problem, reviews spatial-temporal data representations and evaluation metrics, and discusses deep learning baselines and their limitations. Section III gives an overview of the NAS framework, discusses search space design for spatial-temporal traffic data, and reviews NAS search strategies applied to traffic prediction. Section IV discusses open challenges and future research directions. Section V concludes the survey.

\section{Background}\label{sec:background}
\subsection{Traffic Prediction Problem}

Traffic prediction is the task of forecasting future traffic conditions based on past observations \cite{li2018dcrnn}. The quantities of interest include traffic speed (how fast vehicles move), flow (how many vehicles pass a point per unit time), and occupancy (the fraction of time a detector is occupied) \cite{li2018dcrnn}, as well as travel demand such as origin-destination trip volumes \cite{yao2018deepmvst}. These quantities are collected by fixed road sensors, loop detectors, GPS probes, or ride-hailing platforms. Predicting them accurately supports applications, such as adaptive traffic signal control \cite{chavhan2020prediction}, route guidance and navigation \cite{dai2020hstgcn}, and on-demand ride-hailing dispatch \cite{yao2018deepmvst}.

Traffic data has a spatial-temporal structure that shapes how prediction models are built. Road sensors are placed on networks that form irregular graphs, represented as $\mathcal{G} = (\mathcal{V}, \mathcal{E}, \mathbf{A})$, where $\mathcal{V}$ is the set of $N$ sensor nodes, $\mathcal{E}$ is the set of road segments connecting them, and $\mathbf{A} \in \mathbb{R}^{N \times N}$ is an adjacency matrix encoding spatial relationships \cite{yu2018stgcn}. The adjacency matrix can be constructed in different ways: a distance-based matrix sets $A_{ij}$ according to road distance between sensors, a binary connectivity matrix marks directly connected segments, and an adaptive or learned matrix is trained from data to capture functional relationships that go beyond physical proximity \cite{shao2022decoupled}. Temporally, each node records $F$ traffic features at every time step, giving a feature matrix $\mathbf{X}_t \in \mathbb{R}^{N \times F}$ at time $t$. The standard prediction task is: given $T$ historical observations $\{\mathbf{X}_{t-T+1}, \ldots, \mathbf{X}_t\}$, predict the next $H$ steps $\{\mathbf{X}_{t+1}, \ldots, \mathbf{X}_{t+H}\}$ \cite{pan2021autostg}.

Model performance is evaluated using three standard metrics: Mean Absolute Error (MAE), Root Mean Squared Error (RMSE), and Mean Absolute Percentage Error (MAPE), on benchmarks such as METR-LA (207 sensors on Los Angeles highways), PeMS-BAY (325 sensors in the San Francisco Bay Area), and PeMS04/08 from the California PeMS system \cite{li2018dcrnn}. RMSE is more sensitive to large errors, while MAPE measures relative error and is useful when comparing across sensors with different traffic volumes. More recent work uses LargeST \cite{liu2023largest}, which covers over 8,600 sensors and tests model scalability on a much larger network.

Traffic prediction tasks also differ in their spatial and temporal granularity. Predictions can target individual road segments, sensor stations, or aggregated geographic zones \cite{zhang2017stresnet}. Temporally, the standard multi-step setting predicts 12 time steps ahead at 5-minute intervals; errors accumulate across the horizon and architecture choices that work well for short-range prediction may not be optimal for long-range prediction \cite{lyu2025autostf}. This variability in task setup means that a model fine-tuned for one granularity or horizon must often be redesigned for a different setting, which further motivates automated architecture search.

\subsection{Current Challenges in Traffic Prediction}

Traffic prediction is hard because road traffic involves both complex spatial and temporal dependencies that are tightly coupled \cite{li2018dcrnn}. Traffic at one location is influenced by nearby and distant roads, and these interactions play out over time, creating coupled spatial-temporal effects. Road networks also form irregular graphs where sensors are not on a regular grid, so standard convolution operations cannot be applied directly; models instead use graph-based representations that capture the connectivity structure of any road network \cite{yu2018stgcn}.

Traffic patterns are also non-stationary. Traffic data shows strong periodic structure at multiple time scales: daily rush-hour peaks, weekday versus weekend differences, and seasonal shifts driven by weather or school calendars \cite{liu2023largest}. Deep learning models can learn these regularities by including time-of-day and day-of-week embeddings as additional input features \cite{jiang2023pdformer}. However, unexpected events such as accidents, road closures, or large public gatherings break these patterns in ways that are difficult to foresee \cite{pan2021autostg}. Data quality adds a further complication: in real deployments, sensor failure rates can be significant, and missing or corrupted readings must be handled carefully to avoid degrading model inputs \cite{liu2023largest}. Self-supervised training strategies that use masking-and-reconstruction objectives \cite{ji2023stssl} and multi-modal approaches that fuse complementary data sources \cite{deng2024mossl} have been proposed to improve robustness, but neither eliminates the fundamental challenge of sudden pattern shifts.

Two structural challenges remain largely open. First, scalability: many graph-based models have quadratic complexity in the number of nodes, which becomes prohibitive on large networks like LargeST \cite{liu2023largest} with over 8,600 sensors. Recent work explores sparse graph operations and patch-based hierarchical designs to improve this \cite{fang2025patchstg, jiang2024sagdfn}. Second, generalization: a model trained on one city often fails on another because the graph topology, traffic patterns, and sensor density differ \cite{pan2021autostg}. Transfer learning \cite{huang2023traffictl} and federated approaches \cite{zhang2024pfedctp} have been proposed, but the problem is not fully solved. These challenges reveal a structural limit of hand-designed models: every architectural choice is fixed at design time, so a model has no systematic way to adapt when the network size, city layout, or traffic regime changes. NAS addresses this directly, rather than tuning one architecture per dataset by hand, it runs a principled search each time the deployment context shifts, and its effectiveness depends on how well that search handles the specific scalability and generalization pressures of traffic data \cite{lyu2025autostf}.

\subsection{Traditional and Deep Learning Approaches}

Early traffic prediction methods used statistical time series models such as ARIMA and its variants, and Kalman filter-based approaches. These work reasonably well on small, stationary datasets but cannot capture nonlinear spatial-temporal patterns or scale to large sensor networks \cite{yu2018stgcn}.

Deep learning models have largely replaced statistical approaches by learning directly from raw sensor data \cite{li2018dcrnn}. Recurrent neural networks (RNN), which includes Long Short-Term Memory (LSTM) and Gated Recurrent Unit (GRU) networks, model the temporal sequence at each sensor but treat sensors independently, missing spatial structure \cite{li2018dcrnn}. Convolutional networks (CNN) capture local spatial patterns in grid-based representations but cannot handle the irregular graph structure of real road networks \cite{zhang2017stresnet}. GCN and broader Graph Neural Networks (GNN) address this by operating directly on road network graphs and have become the dominant spatial modeling component in traffic prediction \cite{yu2018stgcn}. Transformer-based models extend this with attention mechanisms that capture long-range dependencies across both space and time \cite{jiang2023pdformer}. The best-performing models combine spatial and temporal modules in a single hybrid architecture. For example, Diffusion Convolutional Recurrent Neural Network (DCRNN) couples diffusion-based graph convolution with gated recurrent units, while Spatio-Temporal Graph Convolutional Network (STGCN) alternates graph convolution with temporal convolution, demonstrating that jointly modeling spatial and temporal patterns is essential for accurate traffic prediction \cite{shao2022decoupled}.

More recently, new training paradigms have extended what these architectures can do. Self-supervised learning \cite{ji2023stssl, deng2024mossl} pre-trains models on unlabeled traffic data using augmentation and pretext-task objectives, learning general spatial-temporal representations that can be fine-tuned for specific forecasting tasks. Diffusion-based generative models \cite{wen2023diffstg} capture uncertainty in predictions and produce probabilistic forecasts rather than single point estimates. At the largest scale, spatial-temporal foundation models \cite{yuan2024unist, li2024opencity} are pre-trained on data from multiple cities and aim to generalize across regions with little or no retraining. While these directions push performance further, they also enlarge the space of possible architectural designs, making it harder than ever to choose the right model manually and strengthening the case for automated architecture search \cite{lyu2025autostf}.

Despite their strong results, all of these models are designed by hand. Every architectural choice, which graph convolution to use, how many layers, how to combine spatial and temporal components, must be decided by a human expert, and the best design often changes from one dataset to another \cite{li2020autost, pan2021autostg, wu2022autocts}. This manual process is slow, requires deep domain knowledge, and produces architectures that are tuned to specific benchmarks rather than general traffic prediction \cite{wu2022autocts}. This limitation directly motivates the use of NAS to automate these design decisions.

\section{NAS for Traffic Prediction}\label{sec:nas}

\subsection{NAS Framework Overview}

NAS is the branch of AutoML that automates neural network architecture design. In traditional deep learning, an expert decides every structural detail of a model by hand. They choose how many layers the network has, what operation each layer performs, such as convolution, attention, recurrent units, and how these layers are wired together. Getting these choices right is not straightforward. A researcher usually starts with a known architecture, makes small changes, trains the model, checks the results, and repeats. This back-and-forth can take weeks, and even then the result is often a model that works well on the dataset it was tested on but needs to be redesigned from scratch when the task or data changes \cite{elsken2019nas}.

NAS replaces this manual process with a systematic search. Given a predefined set of candidate operations and connection patterns, (the search space), a NAS algorithm explores possible architectures automatically and selects the one that performs best on a validation set, without requiring a human to evaluate each candidate by hand \cite{wang2024advances}. Because NAS can explore a far larger design space than any individual could evaluate manually, studies across image classification, language modeling, and time series forecasting have shown that NAS-found architectures often match or outperform carefully hand-designed ones \cite{white2023nas1000}. For traffic prediction, this is especially valuable: the best combination of spatial and temporal operations changes from one city to another and from one dataset to another, making manual search impractical \cite{lyu2025autostf}.

Formally, NAS solves a bi-level optimization problem \cite{yang2022iotautoml}:
\begin{equation}
\begin{cases}
f^* = \underset{f \in \mathcal{F}}{\arg\min}\ \mathcal{C}(f(\theta^*),\, D_{\text{val}}) \\
\theta^* = \underset{\theta}{\arg\min}\ \mathcal{L}(f(\theta),\, D_{\text{train}})
\end{cases}
\end{equation}
where $\mathcal{F}$ is the search space of candidate architectures, $f$ is a neural network, $\theta$ are its weights, $D_{\text{train}}$ and $D_{\text{val}}$ are the training and validation sets, $\mathcal{L}$ is the training loss, and $\mathcal{C}$ is the evaluation metric (e.g., MAE or RMSE for traffic prediction). The inner problem optimizes weights on training data. The outer problem then selects the architecture with the best validation performance.

Every NAS method has three core components. The \textit{search space} defines which operations (e.g., convolution, attention, recurrent units) are available and how they can be connected. The \textit{search strategy} determines how to explore this space, examples include reinforcement learning, evolutionary algorithms, gradient-based optimization, and Bayesian optimization. The \textit{performance estimation strategy} evaluates candidate architectures without training each one from scratch, which would be too costly; common techniques include weight sharing across candidates, early stopping, and zero-cost proxy metrics \cite{liu2019darts}. Table~\ref{tab:nas_components} summarizes these three components and their application to traffic prediction.

\begin{table}[t]
\centering
\caption{The three core components of a NAS framework and their role in traffic prediction.}
\label{tab:nas_components}
\renewcommand{\arraystretch}{1.3}
\begin{tabular}{p{1.5cm} p{2.1cm} p{3.5cm}}
\hline
\textbf{Component} & \textbf{Role} & \textbf{Traffic NAS Examples} \\
\hline
Search Space & Defines candidate operations and connections & Spatial ops: graph conv, diffusion, adaptive adjacency; Temporal ops: LSTM, Temporal Convolutional Network (TCN), attention \cite{wu2022autocts} \\
\hline
Search Strategy & Determines how to explore the space & Gradient-based \cite{pan2021autostg}, evolutionary \cite{klosa2023lenas}, one-shot \cite{lyu2025autostf} \\
\hline
Performance Estimation & Evaluates architectures without full training & Weight sharing via supernet \cite{wu2022autocts}, early stopping, zero-cost proxies \\
\hline
\end{tabular}
\end{table}

For traffic prediction, each of these components must be adapted to spatial-temporal data. The search space must cover both spatial and temporal operations. Spatial operations handle the graph structure of road networks, and temporal operations handle the sequential nature of sensor readings. The search strategy must be efficient enough for graph-structured data, where a single architecture evaluation is more expensive than on standard image or language benchmarks because graph convolutions scale with the number of edges. Performance estimation is harder because traffic data is non-stationary: a model that performs well on one subset of the data may perform poorly on a different time period or a different city, so proxy evaluations must be chosen carefully to reflect real deployment conditions \cite{li2020autost}.

The three NAS components map directly to the challenges identified in Section~\ref{sec:background}. The irregular graph structure of road networks is addressed through spatial operation candidates in the search space, including learnable adjacency matrices that adapt the graph to each dataset \cite{pan2021autostg}. The multi-step prediction difficulty is handled by searching over temporal operation types and dilation rates, finding combinations that balance short-range and long-range temporal coverage \cite{wu2022autocts}. The need for generalization across cities is addressed, partially, by searching for architectures that do not overfit to a single graph structure, though as discussed in Section~\ref{sec:challenges}, full cross-city generalization through NAS remains an open problem.

\subsection{Search Space Design for Spatial-Temporal Traffic Data}

The search space is the most critical design decision in NAS: it determines what kinds of architectures can be found, and a poorly defined space will miss the best solutions regardless of the search strategy \cite{liu2019darts}. Three main types of search spaces are used. A \textit{macro} search space treats the entire architecture as a sequence of operation choices and searches over all positions at once; it is flexible but makes the search problem large. A \textit{cell-based} space, introduced by Differentiable Architecture Search (DARTS) \cite{liu2019darts}, instead searches for a small repeatable computational cell that is then stacked to form the full network, reducing search complexity while still allowing expressive architectures. A \textit{hierarchical} space combines both levels, searching over operations within cells and over how cells are connected at the network level \cite{deng2025hierarchicalnas}.

For spatial modeling, candidate operations in traffic NAS include spectral graph convolutions, spatial graph attention, and learnable adjacency matrix operations in which the graph structure is itself part of the search \cite{pan2021autostg, ke2023autostgplus}. Learnable adjacency is especially important for traffic prediction because fixed graph structures based on road distances can miss functional relationships between distant but behaviorally correlated locations \cite{shao2022decoupled}. Some methods also include identity and zero operations as candidates, which allows the search to prune unnecessary spatial connections.

For temporal modeling, typical candidates include LSTM and GRU cells for long-range sequential dependencies, dilated causal convolutions for efficient local pattern capture, and multi-head attention for modeling arbitrary-distance temporal relationships \cite{li2020autost, wu2022autocts}. The interaction between spatial and temporal choices is important: different combinations capture different kinds of spatial-temporal coupling, and the best combination varies across datasets.

The size of the search space directly affects how hard the search problem is. If the spatial candidate set has $M_s$ operations and the temporal candidate set has $M_t$ operations, and the architecture stacks $L$ independent layers, the total space grows as $(M_s \times M_t)^L$ \cite{pan2021autostg}. In practice, traffic NAS methods constrain the space through cell sharing, weight tying, or limiting the search depth to keep it tractable. Multi-scale temporal modeling is an especially important design dimension: including candidate operations at different dilation rates (e.g., dilations of 1, 2, 4, 8) allows the model to capture both fine-grained short-term fluctuations and coarser daily or weekly trends within the same cell \cite{wu2022autocts, lyu2025autostf}. Choosing the right set of dilation rates is itself a design decision that NAS can automate.

Most early methods define a joint search space where spatial and temporal operations are searched together within a shared cell \cite{li2020autost, pan2021autostg}. A more recent approach is \textit{decoupled} search, where spatial and temporal subspaces are searched independently and the best components are assembled afterward \cite{lyu2025autostf}. Decoupling reduces the size of the search space for each stage, lowering memory cost and search time while preserving the quality of the final architecture.

\subsection{NAS Search Strategies Applied to Traffic Prediction}

The search strategy is the algorithm that decides which candidate architectures to evaluate and in what order. It is the core algorithmic component of NAS, and the choice of strategy determines how the total search cost is distributed across architecture evaluations \cite{elsken2019nas}. Four main families exist. \textit{Gradient-based} methods relax the discrete choice of operation to a continuous mixture, so architecture parameters can be updated by gradient descent alongside network weights in a single training run \cite{liu2019darts}. \textit{Evolutionary} methods maintain a population of candidate architectures and improve it over generations through mutation and selection; they require no gradients and avoid the instability problems that affect gradient-based search, but need many evaluations to converge \cite{elsken2019nas}. \textit{One-shot and weight-sharing} methods train a single supernet whose weights are shared across all candidate architectures; any architecture can then be evaluated by sampling its corresponding subgraph without additional training, which greatly lowers search cost \cite{white2023nas1000}. \textit{Reinforcement learning (RL) and Bayesian optimization} treat architecture selection as a sequential decision problem or model the performance landscape probabilistically; both avoid gradient bias but are expensive because each architecture must be evaluated separately before the controller or surrogate can improve \cite{zoph2017nas}. Each strategy makes a different trade-off between search cost, architecture quality, and robustness to search instability, and these trade-offs play out differently on graph-structured traffic data than on standard image benchmarks.

Table~\ref{tab:nas_methods} summarizes the main NAS methods for traffic prediction. Because methods are evaluated on different datasets, prediction targets, and metric scales, MAE values are not directly comparable across all entries; the table should be read as a contextual summary rather than a strict ranking. We discuss each strategy below.

\begin{table*}[t]
\centering
\caption{Summary of NAS methods for traffic prediction, with primary benchmark, MAE, and approximate search cost (GPU-hours).}
\label{tab:nas_methods}
\renewcommand{\arraystretch}{1.2}
\begin{tabular}{p{1.5cm} c p{1.5cm} p{1.6cm} c c p{4.5cm}}
\hline
\textbf{Method} & \textbf{Year} & \textbf{Strategy} & \textbf{Dataset} & \textbf{MAE} & \textbf{Search (h)} & \textbf{Key Contribution} \\
\hline
AutoST \cite{li2020autost}       & 2020 & Gradient    & TaxiBJ$^a$        & —         & $\sim$5       & First DARTS-style Spatio-Temporal (ST) NAS; macro space \\
AutoSTG \cite{pan2021autostg}    & 2021 & Gradient    & METR-LA           & 3.02      & $\sim$10      & Learnable graph structure + operation search \\
AutoCTS \cite{wu2022autocts}     & 2022 & One-shot    & METR-LA           & 3.47      & 21            & Supernet for correlated time series \\
Klosa et al. \cite{klosa2022evolutionary} & 2022 & Evolutionary & METR-LA & 3.22  & $\sim$1200$^b$ & Genetic search for traffic architectures \\
AutoSTG+ \cite{ke2023autostgplus} & 2023 & Gradient   & METR-LA           & 3.00      & $\sim$12      & Improved stability; multi-dataset evaluation \\
AutoCTS+ \cite{wu2023autoctsplus} & 2023 & One-shot   & PeMS03$^c$        & 14.60     & $\sim$190     & Joint architecture + hyperparameter search \\
AutoSTL \cite{zhang2023autostl}   & 2023 & Gradient   & PeMS04$^d$        & —         & $\sim$2       & Joint spatial-temporal multi-task learning \\
NASST \cite{zhang2023nasst}       & 2023 & Gradient   & NYC Bike$^e$      & 1.70      & N/R           & Differentiable ST search with CA fusion \\
LENAS \cite{klosa2023lenas}       & 2023 & Evolutionary & METR-LA         & 4.00      & 1--4          & Low-cost evolutionary search with proxies \\
AutoSTF \cite{lyu2025autostf}     & 2025 & One-shot   & METR-LA           & 3.35      & $\sim$25$^f$  & Decoupled spatial-temporal search \\
\hline
\multicolumn{7}{l}{\footnotesize $^a$ AutoST uses city-flow grid data; MAE units differ from sensor speed benchmarks and are not directly comparable.} \\
\multicolumn{7}{l}{\footnotesize $^b$ Estimated from $\approx$50 GPU-days reported; exact cost depends on hardware.} \\
\multicolumn{7}{l}{\footnotesize $^c$ AutoCTS+ does not evaluate on METR-LA; best MAE on PeMS03 (P-12/Q-12 setting) is shown.} \\
\multicolumn{7}{l}{\footnotesize $^d$ AutoSTL targets multi-task prediction (flow + speed jointly); per-task MAE is not directly comparable across methods.} \\
\multicolumn{7}{l}{\footnotesize $^e$ NASST uses NYC Citi Bike demand data; MAE units differ from highway speed benchmarks. N/R = not reported.} \\
\multicolumn{7}{l}{\footnotesize $^f$ Estimated from per-epoch training time reported in the paper.} \\
\end{tabular}
\end{table*}

\subsubsection{Gradient-based NAS}
DARTS \cite{liu2019darts} relaxes discrete architecture selection to a continuous mixture. At each position in the network, instead of committing to one operation, a \textit{mixed operation} computes a weighted sum of all candidate operation outputs: $\bar{o}(x) = \sum_{o \in \mathcal{O}} \frac{\exp(\alpha_o)}{\sum_{o'} \exp(\alpha_{o'})} \cdot o(x)$, where $\alpha_o$ is a learnable architecture parameter for operation $o$. Both the architecture parameters $\alpha$ and the network weights $w$ are updated jointly by gradient descent — $\alpha$ on a validation set and $w$ on a training set. At the end of search, the operation with the highest $\alpha$ at each position is selected. This approach makes the search fast, a single bilevel training run covers the entire space, but requires all candidate operations to be active in memory simultaneously, which is costly on large graphs \cite{lyu2025autostf}.

AutoST \cite{li2020autost} was the first to apply this approach to spatial-temporal prediction, defining a macro search space with spatial and temporal candidates. AutoSTG \cite{pan2021autostg} extended the idea to graph-structured traffic data by simultaneously searching over the graph adjacency matrix and the computation operations, using a meta-graph formulation. AutoSTG+ \cite{ke2023autostgplus} improved AutoSTG with more stable training and stronger results across multiple benchmarks. AutoSTL \cite{zhang2023autostl} introduced joint spatial-temporal learning patterns to handle multiple correlated traffic signals, while NASST \cite{zhang2023nasst} focused on search efficiency through a more compact differentiable formulation. A common weakness of gradient-based NAS is architecture collapse, where the search over-selects skip connections and produces degenerate architectures; remedies such as random-feature regularization \cite{zhang2023rfdarts} and zero-cost operation scoring \cite{xiang2023zerocostdarts} have been proposed for image tasks but remain underexplored in the traffic domain.

\subsubsection{Evolutionary NAS}
Evolutionary methods maintain a population of candidate architectures. This population is evolved over generations through mutation and selection \cite{klosa2022evolutionary}. Klosa et al. \cite{klosa2022evolutionary} applied this to traffic forecasting, encoding architectures as genomes and using validation accuracy on traffic benchmarks as the fitness signal. LENAS \cite{klosa2023lenas} reduced the cost of this process through proxy evaluations: candidates are first assessed on a small data subset or for fewer training steps, and only the most promising ones are fully trained. Evolutionary search avoids the gradient-bias issues of DARTS-style methods but can still require many evaluations to converge on large search spaces.

\subsubsection{One-shot and Weight-sharing NAS}
One-shot NAS trains a single supernet that encodes all candidate architectures as subgraphs sharing the same weights \cite{wu2022autocts}. Once the supernet is trained, any architecture can be evaluated by sampling its corresponding subgraph without additional training, which greatly reduces the total search cost.

AutoCTS \cite{wu2022autocts} applied this strategy to correlated time series forecasting, building a supernet over spatial and temporal operation candidates and using an evolutionary controller to select the best subgraph. AutoCTS+ \cite{wu2023autoctsplus} extended this with joint search over both the architecture and hyperparameters such as number of layers and learning rate. This automates more of the model development pipeline. AutoSTF \cite{lyu2025autostf} further reduced search cost with a decoupled supernet design: spatial and temporal components are searched in separate, smaller supernets and then combined, lowering memory requirements while maintaining competitive accuracy on multiple traffic benchmarks. A known limitation of weight-sharing approaches is the \textit{coupling problem}: because all sub-architectures share the same weights during supernet training, the shared weights are a compromise that may not accurately reflect any single architecture's true performance when evaluated in isolation \cite{lyu2025autostf}. Decoupled search partially mitigates this by reducing the number of architectures competing for the same weights in each sub-supernet.

\subsubsection{RL-based and Bayesian Optimization}
Reinforcement learning NAS trains a controller to sequentially sample architecture decisions, using validation performance as the reward signal \cite{zoph2017nas}. Bayesian optimization builds a probabilistic surrogate of the performance landscape and picks the next candidate by balancing exploration with exploitation \cite{yang2022iotautoml}. Both strategies can in principle avoid the gradient-bias issues of DARTS-style relaxation and impose no memory cost from holding all candidate operations active at once. However, both require a large number of separate architecture evaluations before converging to a good design \cite{elsken2019nas}. On graph-structured traffic data, each evaluation is substantially more expensive than on image benchmarks because graph convolutions scale with node and edge count, making a single training run on METR-LA or PeMS-BAY take hours even for compact models. The cost-per-evaluation penalty erases the sample-efficiency advantage that RL and Bayesian methods hold over random search, which may explain the limited number of traffic-specific studies using these strategies. The most direct path to making them practical is pairing them with cheap proxy evaluations — as LENAS \cite{klosa2023lenas} demonstrated for evolutionary search — so the controller or surrogate receives fast approximate signals rather than waiting for full training runs.

Across the three main strategy families, a common pattern emerges: each reduces search cost compared to evaluating architectures one by one, but each also introduces trade-offs that are sharpened by the properties of spatial-temporal traffic data. Gradient-based methods are fast but memory-intensive and prone to architecture collapse on large graphs. Evolutionary methods avoid gradient issues but require many evaluations, which is expensive when each training run involves graph convolutions over hundreds of sensors. One-shot methods offer the best cost-quality balance on current benchmarks but suffer from weight-coupling artifacts that can mislead the search. RL-based and Bayesian methods remain largely unexplored in the traffic domain due to their high per-evaluation cost. Taken together, the search space designs and the strategies reviewed here show that NAS can find competitive traffic prediction architectures, but several structural problems remain open.

\section{Open Challenges and Future Directions}\label{sec:challenges}
Despite recent progress, several NAS-specific limitations must be addressed before these methods can be applied at real-world scale.

\subsubsection{Computational Cost}
Even efficient gradient-based methods require hundreds of GPU-hours on standard benchmarks \cite{wu2022autocts, klosa2022evolutionary}, and this grows quickly for large networks. The LargeST benchmark \cite{liu2023largest}, which covers over 8,000 sensors, shows that current NAS pipelines are impractical at city scale. The most promising NAS-specific path forward is zero-cost proxy scoring \cite{xiang2023zerocostdarts}, which ranks candidate architectures using gradient statistics at initialization rather than training each one, cutting the cost of each evaluation from hours to seconds. Combining zero-cost proxies with one-shot supernets \cite{lyu2025autostf} could make NAS practical on large road networks without sacrificing the quality of the found architecture.

\subsubsection{Manual Search Space Design}
In every existing method, a human expert decides which spatial and temporal operations to include as candidates \cite{white2023nas1000}. If the right operation is missing, NAS cannot find it, so the quality of the result is bounded by the designer's prior knowledge \cite{pan2021autostg, lyu2025autostf}. A more systematic approach would use performance data from prior NAS runs across multiple traffic datasets to identify which operations tend to matter and which can be safely excluded, effectively learning a smaller, better-calibrated search space from evidence rather than intuition. Until such methods exist, NAS shifts rather than eliminates the need for domain expertise.

\subsubsection{Cross-City Transferability}
Architectures found by NAS on one dataset do not transfer well to new cities. Current methods search for a fixed architecture on a single benchmark, so moving to a city with a different graph topology or sensor density requires re-running the full search \cite{pan2021autostg, ke2023autostgplus}. The NAS-specific insight here is that the search itself should be designed to produce transferable results. One concrete direction is to meta-learn the search space parameters across cities \cite{huang2023traffictl}, so that a new search starts from an initialization already biased toward operations that work broadly across traffic networks, rather than from scratch each time.

\subsubsection{Fixed Graph Structure}
Most NAS methods assume a fixed graph structure, but real road networks change dynamically due to accidents, construction, and demand shifts \cite{shao2022decoupled}. Current methods include static learned adjacency as a search option \cite{pan2021autostg}, but this adjacency is fixed at training time and cannot respond to real-time changes. A natural extension is to include \textit{graph update rules}, operations that revise the adjacency matrix at each step based on current traffic state, as candidates in the search space. Searching over how the graph evolves, not just what operations run on it, would let NAS find architectures that adapt to dynamic network conditions.

\subsubsection{NAS for Spatial-Temporal Foundation Models}
The rise of spatial-temporal foundation models such as UniST \cite{yuan2024unist} and OpenCity \cite{li2024opencity}, which are pre-trained on multi-city data, raises a new question for NAS: should the search space include pre-trained spatial-temporal modules as candidate operations? A foundation-model-aware NAS could search for how to compose or fine-tune these modules for a target city rather than building architectures from scratch, potentially combining the generalization of pre-training with the adaptability of architecture search.

\begin{table}[t]
\centering
\caption{Summary of open challenges in NAS for traffic prediction.}
\label{tab:challenges}
\renewcommand{\arraystretch}{1.3}
\begin{tabular}{p{2.0cm} p{5.5cm}}
\hline
\textbf{Challenge} & \textbf{Description} \\
\hline
Computational Cost \cite{wu2022autocts, klosa2022evolutionary} & Current NAS pipelines require hundreds of GPU-hours even on small benchmarks, making them impractical on large road networks. \\
\hline
Manual Search Space Design \cite{white2023nas1000} & Every existing method relies on a human expert to select candidate operations, so the quality of the found architecture is bounded by prior knowledge. \\
\hline
Cross-City Transferability \cite{huang2023traffictl} & Architectures found on one dataset do not generalize to cities with different graph topologies or sensor densities, requiring a full re-search each time. \\
\hline
Fixed Graph Structure \cite{shao2022decoupled} & Most NAS methods treat the road graph as static, and cannot adapt to real-time changes caused by accidents, construction, or demand shifts. \\
\hline
NAS for Foundation Models \cite{yuan2024unist, li2024opencity} & It remains unclear how NAS should be redesigned when strong pre-trained spatial-temporal modules are available as candidate operations. \\
\hline
\end{tabular}
\end{table}

\section{Conclusion}\label{Conclusion}
This survey reviewed NAS methods for traffic prediction, organized by search strategy. Gradient-based methods such as AutoST and AutoSTG were the first to apply differentiable search to spatial-temporal data. Evolutionary methods such as LENAS offer a gradient-free alternative at lower per-run cost. One-shot methods such as AutoCTS and AutoSTF train a shared supernet to reduce total search time, and currently offer the best balance of cost and accuracy on standard benchmarks. Across all strategies, NAS-found architectures frequently match or outperform hand-designed baselines on their respective benchmarks, suggesting that automated search can add value over manual design.

Despite this progress, five open challenges remain. First, current NAS pipelines are too expensive for large road networks; zero-cost proxy evaluations offer the most direct path to making them tractable. Second, architectures found on one benchmark rarely transfer to new cities, pointing to the need for cross-city search protocols. Third, search spaces are still designed by hand, so the quality of results depends on the designer's prior knowledge. Fourth, most methods assume a fixed graph structure and cannot adapt to real-time changes in road networks. Finally, the rise of spatial-temporal foundation models raises a new question: when strong pre-trained modules are available, NAS should shift from searching raw operations to searching how to compose and adapt these modules, a direction that could improve both scalability and generalization.

\section*{Acknowledgment}

This work is partially sponsored by the Natural Sciences and Engineering Research Council of Canada (NSERC) through the Discovery Grants (DG) program.

\bibliographystyle{IEEEtran}
\bibliography{citations}

@IEEEtranBSTCTL{IEEEexample:BSTcontrol,
  CTLdash_repeated_names = {no}
}

@article{yang2022iotautoml,
  title   = {IoT data analytics in dynamic environments: From an automated machine learning perspective},
  author  = {Yang, Li and Shami, Abdallah},
  journal = {Engineering Applications of Artificial Intelligence},
  volume  = {116},
  pages   = {105366},
  year    = {2022},
  doi     = {10.1016/j.engappai.2022.105366}
}

@article{elsken2019nas,
  title   = {Neural Architecture Search: A Survey},
  author  = {Elsken, Thomas and Metzen, Jan Hendrik and Hutter, Frank},
  journal = {Journal of Machine Learning Research},
  volume  = {20},
  pages   = {1--21},
  year    = {2019}
}

@inproceedings{liu2019darts,
  title     = {DARTS: Differentiable Architecture Search},
  author    = {Liu, Hanxiao and Simonyan, Karen and Yang, Yiming},
  booktitle = {International Conference on Learning Representations (ICLR)},
  year      = {2019}
}

@inproceedings{li2020autost,
  title     = {AutoST: Efficient Neural Architecture Search for Spatio-Temporal Prediction},
  author    = {Li, Ting and Zhang, Junbo and Bao, Kainan and Liang, Yuxuan and Li, Yexin and Zheng, Yu},
  booktitle = {Proceedings of the 26th ACM SIGKDD International Conference on Knowledge Discovery and Data Mining (KDD)},
  year      = {2020},
  doi       = {10.1145/3394486.3403122}
}

@inproceedings{pan2021autostg,
  title     = {AutoSTG: Neural Architecture Search for Predictions of Spatio-Temporal Graphs},
  author    = {Pan, Zheyi and Ke, Songyu and Yang, Xiaodu and Liang, Yuxuan and Yu, Yong and Zhang, Junbo and Zheng, Yu},
  booktitle = {Proceedings of the Web Conference 2021 (WWW '21)},
  year      = {2021},
  doi       = {10.1145/3442381.3449816}
}

@article{ke2023autostgplus,
  title   = {AutoSTG+: An automatic framework to discover the optimal network for spatio-temporal graph prediction},
  author  = {Ke, Songyu and Pan, Zheyi and He, Tianfu and Liang, Yuxuan and Zhang, Junbo and Zheng, Yu},
  journal = {Artificial Intelligence},
  volume  = {318},
  pages   = {103899},
  year    = {2023},
  doi     = {10.1016/j.artint.2023.103899}
}

@article{wu2022autocts,
  title   = {AutoCTS: Automated Correlated Time Series Forecasting},
  author  = {Wu, Xinle and Zhang, Dalin and Guo, Chenjuan and He, Chaoyang and Yang, Bin and Jensen, Christian S.},
  journal = {Proceedings of the VLDB Endowment},
  volume  = {15},
  number  = {4},
  pages   = {971--983},
  year    = {2022},
  doi     = {10.14778/3503585.3503604}
}

@article{wu2023autoctsplus,
  title   = {AutoCTS+: Joint Neural Architecture and Hyperparameter Search for Correlated Time Series Forecasting},
  author  = {Wu, Xinle and Zhang, Dalin and Zhang, Miao and Guo, Chenjuan and Yang, Bin and Jensen, Christian S.},
  journal = {Proceedings of the ACM on Management of Data},
  volume  = {1},
  number  = {1},
  pages   = {97:1--97:26},
  year    = {2023},
  doi     = {10.1145/3588951}
}

@inproceedings{klosa2022evolutionary,
  title     = {Evolutionary Neural Architecture Search for Traffic Forecasting},
  author    = {Klosa, Daniel and B{\"u}skens, Christof},
  booktitle = {2022 21st IEEE International Conference on Machine Learning and Applications (ICMLA)},
  pages     = {1230--1237},
  year      = {2022},
  doi       = {10.1109/ICMLA55696.2022.00198}
}

@article{klosa2023lenas,
  title   = {Low Cost Evolutionary Neural Architecture Search ({LENAS}) Applied to Traffic Forecasting},
  author  = {Klosa, Daniel and B{\"u}skens, Christof},
  journal = {Machine Learning and Knowledge Extraction},
  volume  = {5},
  number  = {3},
  pages   = {830--846},
  year    = {2023},
  doi     = {10.3390/make5030044}
}

@inproceedings{lyu2025autostf,
  title     = {AutoSTF: Decoupled Neural Architecture Search for Cost-Effective Automated Spatio-Temporal Forecasting},
  author    = {Lyu, Tengfei and Zhang, Weijia and Deng, Jinliang and Liu, Hao},
  booktitle = {Proceedings of the 31st ACM SIGKDD Conference on Knowledge Discovery and Data Mining (KDD '25)},
  pages     = {985--996},
  year      = {2025},
  doi       = {10.1145/3690624.3709323}
}

@article{jiang2022gnntrafic,
  title   = {Graph Neural Network for Traffic Forecasting: A Survey},
  author  = {Jiang, Weiwei and Luo, Jiayun},
  journal = {Expert Systems with Applications},
  volume  = {207},
  pages   = {117921},
  year    = {2022},
  doi     = {10.1016/j.eswa.2022.117921}
}

@article{shao2022decoupled,
  title   = {Decoupled Dynamic Spatial-Temporal Graph Neural Network for Traffic Forecasting},
  author  = {Shao, Zezhi and Zhang, Zhao and Wei, Wei and Wang, Fei and Xu, Yongjun and Cao, Xin and Jensen, Christian S.},
  journal = {Proceedings of the VLDB Endowment},
  volume  = {15},
  number  = {11},
  pages   = {2733--2746},
  year    = {2022},
  doi     = {10.14778/3551793.3551827}
}

@article{wang2024advances,
  title   = {Advances in Neural Architecture Search},
  author  = {Wang, Xin and Zhu, Wenwu},
  journal = {National Science Review},
  volume  = {11},
  number  = {8},
  pages   = {nwae282},
  year    = {2024},
  doi     = {10.1093/nsr/nwae282}
}

@article{khatiriolyaee2026automl,
  title   = {A Comprehensive Review of Traffic Prediction: From Traditional Machine Learning to AutoML},
  author  = {Khatiriolyaee, Mahshid and Yang, Li and Pazzi, Richard W.},
  journal = {Neurocomputing},
  volume  = {666},
  pages   = {132148},
  year    = {2026},
  doi     = {10.1016/j.neucom.2025.132148}
}

@article{jin2023stgnnsurvey,
  title   = {Spatio-Temporal Graph Neural Networks for Predictive Learning in Urban Computing: A Survey},
  author  = {Jin, Guangyin and Liang, Yuxuan and Fang, Yuchen and Shao, Zezhi and Huang, Jincai and Zhang, Junbo and Zheng, Yu},
  journal = {IEEE Transactions on Knowledge and Data Engineering},
  volume  = {36},
  number  = {10},
  pages   = {5388--5407},
  year    = {2024},
  doi     = {10.1109/TKDE.2023.3333824}
}

@inproceedings{zhang2023autostl,
  title     = {AutoSTL: Automated Spatio-Temporal Multi-Task Learning},
  author    = {Zhang, Zijian and Zhao, Xiangyu and Miao, Hao and Zhang, Chunxu and Zhao, Hongwei and Zhang, Junbo},
  booktitle = {Proceedings of the AAAI Conference on Artificial Intelligence},
  volume    = {37},
  pages     = {4902--4910},
  year      = {2023}
}

@inproceedings{zhang2023nasst,
  title     = {Traffic Spatial-Temporal Prediction Based on Neural Architecture Search},
  author    = {Zhang, Dongran and Luo, Gang and Li, Jun},
  booktitle = {Proceedings of the 18th International Symposium on Spatial and Temporal Data (SSTD '23)},
  pages     = {21--30},
  year      = {2023},
  doi       = {10.1145/3609956.3609962}
}

@article{deng2025hierarchicalnas,
  title     = {Optimizing Time Series Forecasting Architectures: A Hierarchical Neural Architecture Search Approach},
  author    = {Deng, Difan and Lindauer, Marius},
  journal   = {Transactions on Machine Learning Research},
  year      = {2025}
}

@article{white2023nas1000,
  title   = {Neural Architecture Search: Insights from 1000 Papers},
  author  = {White, Colin and Safari, Mahmoud and Sukthanker, Rhea and Ru, Binxin and Elsken, Thomas and Zela, Arber and Dey, Debadeepta and Hutter, Frank},
  journal = {arXiv preprint arXiv:2301.08727},
  year    = {2023}
}

@inproceedings{liu2023largest,
  title     = {LargeST: A Benchmark Dataset for Large-Scale Traffic Forecasting},
  author    = {Liu, Xu and Xia, Yutong and Liang, Yuxuan and Hu, Junfeng and Wang, Yiwei and Bai, Lei and Huang, Chao and Liu, Zhenguang and Hooi, Bryan and Zimmermann, Roger},
  booktitle = {Advances in Neural Information Processing Systems (NeurIPS), Datasets and Benchmarks Track},
  volume    = {36},
  year      = {2023}
}

@inproceedings{jiang2023pdformer,
  title     = {PDFormer: Propagation Delay-Aware Dynamic Long-Range Transformer for Traffic Flow Prediction},
  author    = {Jiang, Jiawei and Han, Chengkai and Zhao, Wayne Xin and Wang, Jingyuan},
  booktitle = {Proceedings of the AAAI Conference on Artificial Intelligence},
  volume    = {37},
  number    = {4},
  pages     = {4365--4373},
  year      = {2023}
}

@inproceedings{fang2025patchstg,
  title     = {Efficient Large-Scale Traffic Forecasting with Transformers: A Spatial Data Management Perspective},
  author    = {Fang, Yuchen and Liang, Yuxuan and Hui, Bo and Shao, Zezhi and Deng, Liwei and Liu, Xu and Jiang, Xinke and Zheng, Kai},
  booktitle = {Proceedings of the 31st ACM SIGKDD Conference on Knowledge Discovery and Data Mining (KDD)},
  year      = {2025}
}

@inproceedings{ji2023stssl,
  title     = {Spatio-Temporal Self-Supervised Learning for Traffic Flow Prediction},
  author    = {Ji, Jiahao and Wang, Jingyuan and Huang, Chao and Wu, Junjie and Xu, Boren and Wu, Zhenhe and Zhang, Junbo and Zheng, Yu},
  booktitle = {Proceedings of the AAAI Conference on Artificial Intelligence},
  volume    = {37},
  number    = {4},
  pages     = {4356--4364},
  year      = {2023}
}

@inproceedings{jiang2024sagdfn,
  title     = {SAGDFN: A Scalable Adaptive Graph Diffusion Forecasting Network for Multivariate Time Series Forecasting},
  author    = {Jiang, Yue and Li, Xiucheng and Chen, Yile and Liu, Shuai and Kong, Weilong and Lentzakis, Antonis F. and Cong, Gao},
  booktitle = {Proceedings of the 40th IEEE International Conference on Data Engineering (ICDE)},
  year      = {2024}
}

@inproceedings{deng2024mossl,
  title     = {Multi-Modality Spatio-Temporal Forecasting via Self-Supervised Learning},
  author    = {Deng, Jiewen and Jiang, Renhe and Zhang, Jiaqi and Song, Xuan},
  booktitle = {Proceedings of the 33rd International Joint Conference on Artificial Intelligence (IJCAI)},
  pages     = {2013--2021},
  year      = {2024}
}

@article{li2024opencity,
  title   = {OpenCity: Open Spatio-Temporal Foundation Models for Traffic Prediction},
  author  = {Li, Zhonghang and Xia, Long and Shi, Lei and Xu, Yong and Yin, Dawei and Huang, Chao},
  journal = {ACM Transactions on Intelligent Systems and Technology},
  year    = {2024},
  doi     = {10.1145/3773912}
}

@inproceedings{yuan2024unist,
  title     = {UniST: A Prompt-Empowered Universal Model for Urban Spatio-Temporal Prediction},
  author    = {Yuan, Yuan and Ding, Jingtao and Feng, Jie and Jin, Depeng and Li, Yong},
  booktitle = {Proceedings of the 30th ACM SIGKDD Conference on Knowledge Discovery and Data Mining},
  pages     = {4095--4106},
  year      = {2024},
  doi       = {10.1145/3637528.3671662}
}

@inproceedings{wen2023diffstg,
  title     = {DiffSTG: Probabilistic Spatio-Temporal Graph Forecasting with Denoising Diffusion Models},
  author    = {Wen, Haomin and Lin, Youfang and Xia, Yutong and Wan, Huaiyu and Wen, Qingsong and Zimmermann, Roger and Liang, Yuxuan},
  booktitle = {Proceedings of the 31st ACM International Conference on Advances in Geographic Information Systems (SIGSPATIAL)},
  year      = {2023}
}

@inproceedings{zhang2023rfdarts,
  title     = {Differentiable Architecture Search with Random Features},
  author    = {Zhang, Xuanyang and Li, Yonggang and Zhang, Xiangyu and Wang, Yongtao and Sun, Jian},
  booktitle = {Proceedings of the IEEE/CVF Conference on Computer Vision and Pattern Recognition (CVPR)},
  year      = {2023}
}

@inproceedings{xiang2023zerocostdarts,
  title     = {Zero-Cost Operation Scoring in Differentiable Architecture Search},
  author    = {Xiang, Lichuan and Dudziak, {\L}ukasz and Abdelfattah, Mohamed S. and Chau, Thomas and Lane, Nicholas D. and Wen, Hongkai},
  booktitle = {Proceedings of the AAAI Conference on Artificial Intelligence},
  volume    = {37},
  number    = {9},
  pages     = {10577--10585},
  year      = {2023}
}

@article{huang2023traffictl,
  title   = {Traffic Prediction with Transfer Learning: A Mutual Information-based Approach},
  author  = {Huang, Yunjie and Song, Xiaozhuang and Zhu, Yuanshao and Zhang, Shiyao and Yu, James J.Q.},
  journal = {IEEE Transactions on Intelligent Transportation Systems},
  volume  = {24},
  number  = {8},
  pages   = {8526--8540},
  year    = {2023},
  doi     = {10.1109/TITS.2023.3266398}
}

@inproceedings{zhang2024pfedctp,
  title     = {Personalized Federated Learning for Cross-City Traffic Prediction},
  author    = {Zhang, Yu and Lu, Hua and Liu, Ning and Xu, Yonghui and Li, Qingzhong and Cui, Lizhen},
  booktitle = {Proceedings of the 33rd International Joint Conference on Artificial Intelligence (IJCAI)},
  pages     = {5526--5534},
  year      = {2024},
  doi       = {10.24963/ijcai.2024/611},
}

@inproceedings{zoph2017nas,
  title     = {Neural Architecture Search with Reinforcement Learning},
  author    = {Zoph, Barret and Le, Quoc V.},
  booktitle = {International Conference on Learning Representations (ICLR)},
  year      = {2017}
}

@inproceedings{li2018dcrnn,
  title     = {Diffusion Convolutional Recurrent Neural Network: Data-Driven Traffic Forecasting},
  author    = {Li, Yaguang and Yu, Rose and Shahabi, Cyrus and Liu, Yan},
  booktitle = {International Conference on Learning Representations (ICLR)},
  year      = {2018}
}

@inproceedings{yu2018stgcn,
  title     = {Spatio-Temporal Graph Convolutional Networks: A Deep Learning Framework for Traffic Forecasting},
  author    = {Yu, Bing and Yin, Haoteng and Zhu, Zhanxing},
  booktitle = {Proceedings of the 27th International Joint Conference on Artificial Intelligence (IJCAI)},
  pages     = {3634--3640},
  year      = {2018}
}

@inproceedings{zhang2017stresnet,
  title     = {Deep Spatio-Temporal Residual Networks for Citywide Crowd Flows Prediction},
  author    = {Zhang, Junbo and Zheng, Yu and Qi, Dekang},
  booktitle = {Proceedings of the Thirty-First AAAI Conference on Artificial Intelligence},
  pages     = {1655--1661},
  year      = {2017}
}

@article{chavhan2020prediction,
  title   = {Prediction Based Traffic Management in a Metropolitan Area},
  author  = {Chavhan, Suresh and Venkataram, Pallapa},
  journal = {Journal of Traffic and Transportation Engineering (English Edition)},
  volume  = {7},
  number  = {4},
  pages   = {447--466},
  year    = {2020},
  doi     = {10.1016/j.jtte.2018.07.004}
}

@inproceedings{dai2020hstgcn,
  title     = {Hybrid Spatio-Temporal Graph Convolutional Network: Improving Traffic Prediction with Navigation Data},
  author    = {Dai, Rui and Xu, Shenkun and Gu, Qian and Ji, Chenguang and Liu, Kaikui},
  booktitle = {Proceedings of the 26th ACM SIGKDD International Conference on Knowledge Discovery and Data Mining (KDD)},
  pages     = {3074--3082},
  year      = {2020},
  doi       = {10.1145/3394486.3403358}
}

@inproceedings{yao2018deepmvst,
  title     = {Deep Multi-View Spatial-Temporal Network for Taxi Demand Prediction},
  author    = {Yao, Huaxiu and Wu, Fei and Ke, Jintao and Tang, Xianfeng and Jia, Yitian and Lu, Siyu and Gong, Pinghua and Ye, Jieping and Li, Zhenhui},
  booktitle = {Proceedings of the Thirty-Second AAAI Conference on Artificial Intelligence},
  pages     = {2588--2595},
  year      = {2018}
}

\vspace{12pt}

\end{document}